\documentclass[letterpaper, 10 pt, journal, twoside]{IEEEtran}

\IEEEoverridecommandlockouts                              

\usepackage{cite}
\usepackage{mathtools} 
\usepackage{amsmath} 
\usepackage{amssymb}  
\usepackage{siunitx}
\usepackage{textcomp}
\usepackage{xcolor}
\usepackage[hidelinks]{hyperref}
\usepackage{cleveref}
\usepackage{subcaption}
\usepackage{graphicx}
\usepackage{booktabs}
\usepackage{framed}
\usepackage{soul}
\usepackage{xcolor}
\usepackage{ulem}
\usepackage{dblfloatfix} 

\usepackage{color}

\newcommand{\fin}[1]{\textcolor{black}{#1}}

\usepackage[font=small,skip=3pt]{caption}

\IEEEaftertitletext{\vspace{-2em}}

\begin{document}

\title{Simple Models, Real Swimming: Digital Twins for Tendon-Driven Underwater Robots}

\author{Mike Y. Michelis$^1$, Nana Obayashi$^{2,3}$, Josie Hughes$^3$, and Robert K. Katzschmann$^1$%
\thanks{$^1$ Soft Robotics Lab and ETH AI Center, ETH Zurich, Switzerland}%
\thanks{$^2$ Prema Lab, New York University, USA}%
\thanks{$^3$ CREATE Lab, EPFL, Switzerland}%
\thanks{\fin{Videos, code, and data are available on our website at: \texttt{\href{srl-ethz.github.io/website_fishsim}{srl-ethz.github.io/website\_fishsim}}}}%
}



\maketitle

\thispagestyle{empty}
\pagestyle{empty}

\begin{abstract}


Mimicking the graceful motion of swimming animals remains a core challenge in soft robotics due to the complexity of fluid-structure interaction and the difficulty of controlling soft, biomimetic bodies. 
Existing modeling approaches are often computationally expensive and impractical for complex control or reinforcement learning needed for realistic motions to emerge in robotic systems. 
In this work, we present a tendon-driven fish robot modeled in an efficient underwater swimmer environment using a simplified, stateless hydrodynamics formulation implemented in the widespread robotics framework MuJoCo. 
With just two real-world swimming trajectories, we identify five fluid parameters that allow a matching to experimental behavior and generalize across a range of actuation frequencies. 
We show that this stateless fluid model can generalize to unseen actuation and outperform classical analytical models such as the elongated body theory.
This simulation environment runs faster than real-time and can easily enable downstream learning algorithms such as reinforcement learning for target tracking, reaching a $93\%$ success rate. Due to the simplicity and ease of use of the model and our open-source simulation environment, our results show that even simple, stateless models --- when carefully matched to physical data --- can serve as effective digital twins for soft underwater robots, opening up new directions for scalable learning and control in aquatic environments.

\end{abstract}



\section{Introduction}


Animals move gracefully and efficiently in their natural habitats, and can perform a broad variety of tasks that present-day robots are still far away from being capable of, from agile locomotion to dexterous manipulation. These animals show controlled actions through a complex interplay between the rigid and soft materials in their bodies with the multiphysical environment around them. 

One such environment that features a very dense and continuous interaction between animal and surrounding is the underwater environment. Unlike robots in locomotion or manipulation, where contact is sparse and discontinuous, underwater settings require a continuous exchange of forces between the robot surface and the surrounding fluid. Through this complex interaction we might find natural and efficient movements emerge in our robot. Several underwater robots have been built to mimic the elegant motions found in nature, such as fish \cite{katzschmann_exploration_2018, youssef_design_2022}, jellyfish \cite{wang_versatile_2023}, and starfish \cite{du_underwater_2021}. However, modeling and control has remained a challenging topic for these soft robotic systems. 

Soft robotics in general has often required computationally expensive modeling approaches in the past, which have to be tuned through system identification \cite{dubied_sim--real_2022, shi2025soft} to match real-world behavior. Simplified modeling has been applied to soft robots that can be approximated as 1D-splines \cite{della_santina_improved_2020} or as rods \cite{zhang2019modeling}, but this severely limits the range of robot geometries that can be considered. Similarly, black-box neural networks can be used to approximate these complex dynamics \cite{zhang_sim2real_2022, lin_dynamic_2024}, but fail to be interpretable or generalize to unseen scenarios.

\begin{figure}[!t]
    \centering
    \includegraphics[width=\columnwidth]{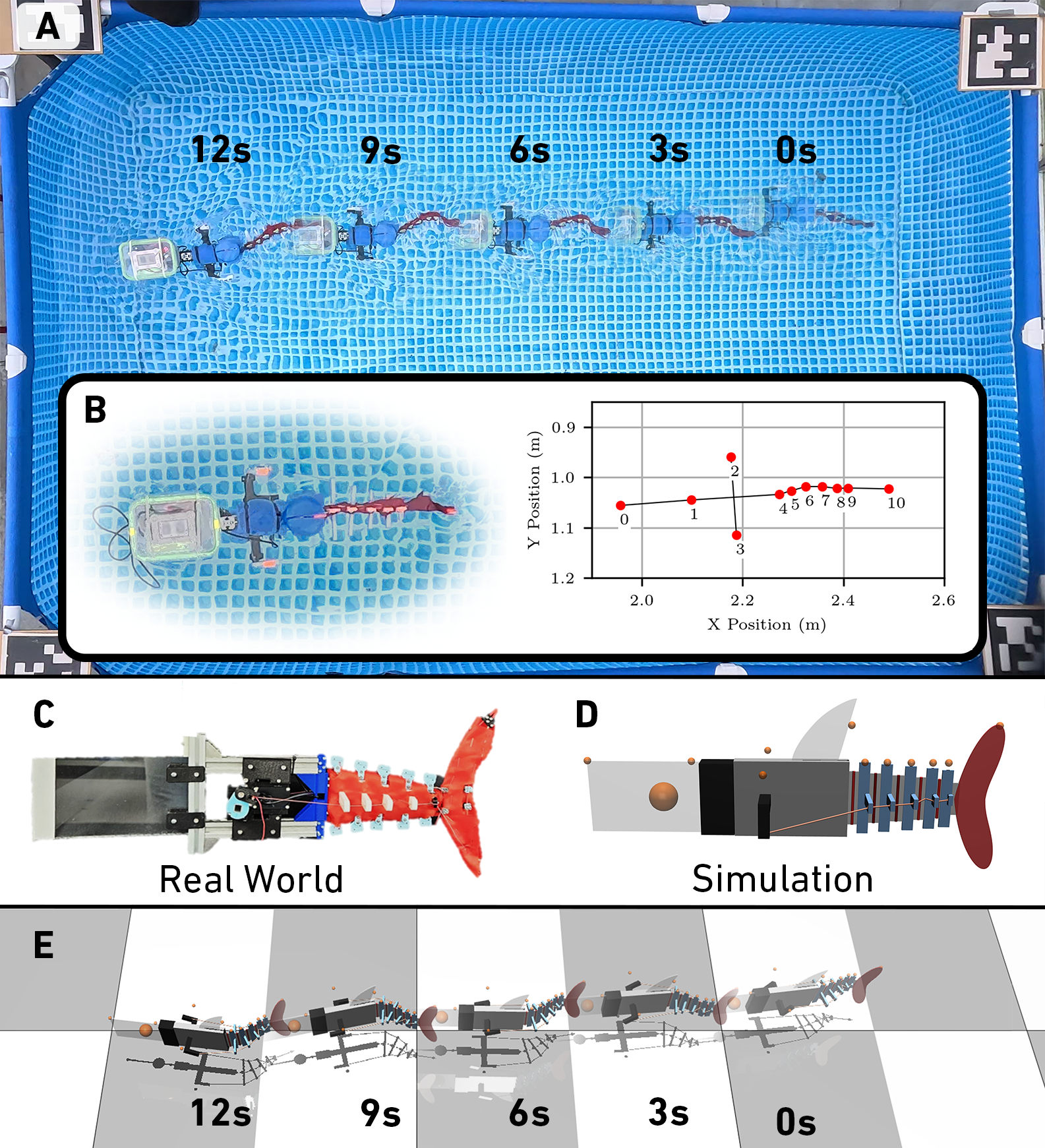}
    \caption{Overview of the simulated and real swimmer robots. \textit{A)} Robot swimming in the pool captured from a top-down view, images overlaid are \SI{3}{\second} apart. \textit{B)} The 11 markers are tracked and extracted from the video. \textit{C)} Side-view of the hardware with a single motor actuating the tendon-driven tail. \textit{D)} Simulated robot with markers indicated by small orange spheres. Center of mass is located at the large orange sphere in the head. \textit{E)} Simulated trajectory after matching fluid model to real experiment.}
    \label{fig:overview}
    \vspace{-0em}
\end{figure}

In this paper we present a computationally efficient approximation to modeling biomimetic tendon-driven swimmers, such that we can further investigate the natural motions that arise from a flexible spine that is actuated to interact with the environment. Previous work \cite{vaxenburg2025whole} has shown a simplified fluid model for analyzing fruit fly behavior. We verify this simplification can match real-world experimental data for underwater swimmers with as little as 2 trajectories, and can also generalize to a wide range of unseen actuation frequencies. We provide the integration of this robot model into standard Reinforcement Learning (RL) pipelines for the broader robot learning community, and we showcase accurate RL agents for target tracking. In summary, we contribute:
\begin{enumerate}
    \item \emph{Simulation environment for underwater tendon-driven robots:} We test a stateless fluid model for underwater swimming.
    
    \item \emph{Sim-to-real matching:} We show the matching and generalization at different constant frequencies of actuation for this simplified fluid model.
    
    \item \emph{Outperforming baseline:} Analytical models such as elongated body theory struggle to describe the swimming of our robot, while our model can generalize to unseen actuation.

    \item \emph{Downstream reinforcement learning:} We demonstrate how our digital twin can be used for specific control tasks, such as learning to track targets.
\end{enumerate}

\section{Related Work}



Computational Fluid Dynamics (CFD) simulations are computationally expensive, but can provide accurate insights in fish swimming behavior in controlled environments \cite{borazjani_numerical_2008}. These accurate simulations can be shown to work in an RL pipeline in 2D \cite{zhang_simulation_2022}, where a surrogate neural network is trained on experimental data such that the RL agent first learns in this faster surrogate environment, then transitions to the accurate CFD when the policy is sufficiently precise. However, even though the training in the surrogate took less than an hour, the finetuning in CFD needed 16 days, hence such an approach remains infeasible to scale to 3D due to computational cost. An efficient approach leveraging GPU parallelization was proposed in \cite{liu_fishgym_2022} using Lattice Boltzmann Method for the fluid and a skinned, articulated rigid skeleton for the fish. This was transferred to a surface swimmer in \cite{lin_learning_2025} with careful tuning of the servo motors and fluid forces. The approach is presented for swimmers with controllable motors in each joint of the tail, whereas in this paper we explore the application to underactuated systems that rely on compliance, such as tendon-driven swimmers with a flexible spine.


The simplest model for fluid dynamics that is solving for a time-dependent fluid state would be particle-based methods such as the material point method \cite{wang2023softzoo}, which can provide qualitatively good results, but struggle especially with boundary conditions. Compared to full CFD, the fluid-structure interaction between robot and fluid has shown to generate physically inaccurate fluid forces for the material point method \cite{lee_unified_2025}, rendering it a poor choice for underwater robotics.

Simplifications can be made to these expensive CFD methods, such as added mass or slender body theorem \cite{lighthill_note_1960}, which consider the local fluid behavior around the submerged object, and lead to stateless thrust/drag formulations with only a few parameters to match to fish swimming experiments \cite{yu_data-driven_2016, du_underwater_2021, chen_experimental_2023, lu_toward_2024}. These methods are stateless since no fluid pressure or velocity field is solved for; the fluid force on the fish is directly computed from the state of the fish and not the state of the fluid. This significantly speeds up the simulation time, but at the cost of not being able to propagate fluid state information over time, ignoring, for example, how fluids interact with wall boundaries or how multiple objects in the same fluid domain interact with one another.

To simulate the flying behavior of a fruit fly, a more expressive fluid model was developed including blunt drag, slender drag, angular drag, Kutta lift, and Magnus lift \cite{vaxenburg2025whole}. These stateless fluidic forces and torques are based on a quasi-steady-state approximation \cite{andersen_analysis_2005} with the added simplification that all bodies experiencing fluid forces are ellipsoids. We will expand this work to underwater robots to show its feasibility in matching real-world swimmer dynamics.

Given enough experimental data, learning-based approaches can be applied to fish swimming as well. A black-box neural network was trained to predict fluid thrust forces of a fixed, pneumatically-actuated sillicone fish tail \cite{zhang_sim2real_2022}, but was shown to struggle while generalizing to the stiffer sillicone tails. Another approach is to learn the fluid dynamics using a physics-informed surrogate model, and combine this into a differentiable simulation pipeline for optimizing the control frequency \cite{nava_fast_2022}. Providing more structure in the learning architecture, \cite{saeed_data-driven_2025} present a data-driven reduced-order model for direct numerical simulations of an undulating swimmer and \cite{lin_dynamic_2024} use a Koopman operator to predict the time-dependent dynamics of a real-world motor-driven fish while accounting for the background flow using a very coarse 2D CFD to reduce computational cost. These applications remain rather task-specific and data-hungry, hence in this paper we propose a more general learning framework that closely follows the learning-based robotics community using simulators such as MuJoCo \cite{todorov_mujoco_2012} for bridging the sim-to-real gap for soft underwater robotics. We demonstrate that our approach can achieve comparable fidelity to learned models, while remaining highly interpretable and efficient, and requiring far less data to calibrate a model matching to reality.

\section{Experimental Setup}

\subsection{Hardware Setup}

The robotic fish design is based on a parametrically length-scalable framework, developed through work \cite{obayashi2025scafi}, which allows the robot to be fabricated at different lengths. The robot's physical frame is split into two distinct sections: the passive front-end and the active, compliant tail. The passive front-end is constructed with acrylic sheets fixed with 3D-printed pieces to aluminum extrusions. Acrylic sheets are also used to construct the pectoral and dorsal fins for stability. A single dual-shaft, waterproof Dynamixel XW-540-T140-R motor is centrally mounted within this front-end. This motor actuates two fishing wire tendons through a crank-slider-inspired mechanism. These tendons run along either side of the tail, with their routing transitioning at the tail's midpoint to create an antagonistic S-bend motion, mimicking aquatic creature kinematics.

The active tail is constructed from four straight \SI{1}{\milli\meter} diameter nitinol rods that run its entire length. These rods provide the structural stiffness and elasticity for propulsion and are rigidly secured to 3D-printed PLA structural supports placed at five discrete, equidistant points along the tail. These supports have guide holes for the actuating tendons, which terminate at the caudal fin. The caudal fin itself is formed by two additional nitinol rods, mounted vertically perpendicular to the tail body with 3D-printed supports. All non-waterproof sensing, computing, and power electronics are housed within a custom waterproof container atop the front-end frame. A 3S LiPo battery powers a DC-DC converter for a Raspberry Pi Zero, and a Dynamixel U2D2 Motor Controller. 
A waterproof resistant fabric covers the tail and part of the body to provide surface area for propulsion, and polyethylene foam is added to make the swimmer buoyant.


The tail of the fish is divided in five spine segments (the PLA structural supports), each of which has a marker placed on top which can be tracked along with the markers on the rigid body, head, and tail fin using a CSRT tracker \cite{lukezic_discriminative_2017} from OpenCV. The bulk size of the robot is roughly \SI{57}{\centi\meter} long, \SI{25}{\centi\meter} tall, and \SI{17}{\centi\meter} wide. We place the swimmer in a \SI{2}{\meter} $\times$ \SI{3}{\meter} pool, capture the camera footage from the top down using a GoPro Hero 8, at a resolution of $2160\times3840$, 60 frames per second, and a linear field of view. This video data can be seen in \Cref{fig:overview}, where the extracted markers are shown as well. The tracked markers are processed by rotating all markers into their local frame where the heading direction of the fish defines the negative x-axis.


\subsection{Simulation Setup}

\begin{figure}[!b]
    \centering
    \includegraphics[width=\columnwidth]{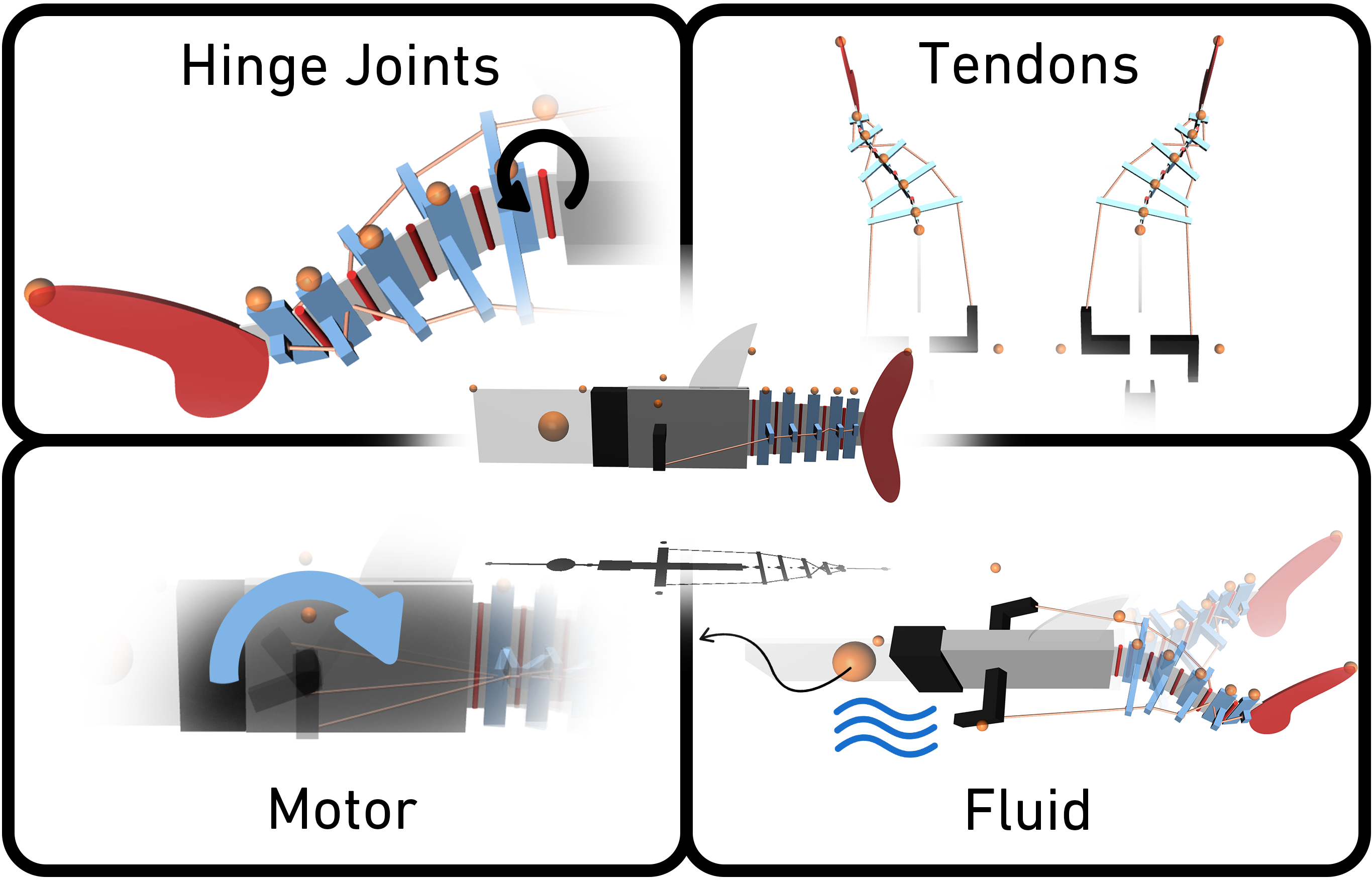}
    \caption{Overview of the main mechanisms for the swimmer simulation environment. We simplify the deformable spine with multiple hinge joints as an articulated rigid body, we use stiff elastic tendons for the tendon-driven actuation, a velocity-controlled motor that pulls the tendons, and a simplified fluid model to mimic real-world swimming behavior.}
    \label{fig:sim}
\end{figure}

We make several simplifications to efficiently simulate the robotic swimmer, these can be seen in \Cref{fig:sim}. Firstly, we assume the continuously bending tail is discretized in rigid segments connected by hinge joints, which has been shown to be a valid approximation for continuum robots \cite{lee_aquarium_2023}. And secondly, a simplified, stateless fluid dynamics model \cite{vaxenburg2025whole} is used to model thrust generation of the fish. This was previously applied on a fruit fly, where we extend its application area to real-time underwater robots as well. Since this fluid dynamics model is implemented in Mujoco \cite{todorov_mujoco_2012}, we implement our robotic fish in this simulation framework as well.

For the articulated rigid body, we choose to discretize based on the spine segments of the hardware setup, as seen in \Cref{fig:overview}. The stiffness of the hinge joints are identified based on experimental data, as explained in \Cref{sec:sysid}, and we use the same stiffness for all tail joints. We model the tendons as stiff springs instead of length constraints, since this results in more computationally stable simulations. From our experiments the length of the tendons stretches roughly \SI{3}{\%} during actuation. Since the buoyant foam on the hardware robot results in a surface swimmer, we mimic this through a soft position constraint in z-direction on the simulated swimmer.

Initially developed for modeling the fluttering of playing cards falling in quiescent and incompressible air, \cite{vaxenburg2025whole} extended this to self-propelling flight of a fruit fly. In this paper, we apply this model to hydrodynamics of a swimmer. The same assumption is made that all bodies that experience the fluid forces are approximated as ellipsoids. The fluid force is defined with the following components:
\begin{align*}
    &\textit{Blunt and Slender Drag:} 
    \\ 
    &\hspace{1em} \vec{f}_D = - \rho \left( c_{blunt} \; A_v + c_{slender} \; \left( A_{max} - A_v \right) \right) \| \vec{v} \| \; \vec{v}
    \\
    &\textit{Angular Drag:} 
    \\ 
    &\hspace{1em} \vec{\tau}_D = - \rho \left( \left( c_{angular} \; \vec{I}_D + c_{slender} \; \left( \vec{I}_{max} - \vec{I}_D \right) \right) \cdot \vec{\omega} \right) \vec{\omega}
    \\
    &\textit{Viscous Drag:} 
    \\
    &\hspace{1em} \vec{f}_V = - 6 \pi \mu \tilde{r} \; \vec{v}
    \\
    &\hspace{1em} \vec{\tau}_V = - 8 \pi \mu \tilde{r}^3 \; \vec{\omega}
    \\
    &\textit{Kutta Lift:} 
    \\ 
    &\hspace{1em} \vec{f}_K = \rho \frac{c_{kutta} \; A_v}{\| \vec{v} \|} \left( \vec{v} \times \vec{v}_{\|} \right) \times \vec{v}
    \\
    &\textit{Magnus Lift:} 
    \\ 
    &\hspace{1em} \vec{f}_M = \rho \; c_{magnus} V \; \vec{\omega} \times \vec{v}
    \\
    &\textit{Added Mass:} 
    \\ 
    &\hspace{1em} \vec{f}_A = - \mathbf{M}_A \dot{\vec{v}} + \left( \mathbf{M}_A \vec{v} \right) \times \vec{\omega}
    \\
    &\hspace{1em} \vec{\tau}_A = - \mathbf{I}_A \dot{\vec{\omega}} + \left( \mathbf{M}_A \vec{v} \right) \times \vec{v} + \left( \mathbf{I}_A \vec{\omega} \right) \times \vec{\omega}
\end{align*}
where $\rho$ is the fluid density, $\vec{v}, \vec{\omega}$ the linear and angular velocity of the body respectively, $A_v$ the projected surface of the ellipsoid normal to the flow, $A_{max}$ the maximum projected surface area, $\vec{I}_D$ the diagonal entries of the moment of inertia of the ellipsoid, $\vec{I}_{max}$ a vector with each component equal to the maximum entry in the moment of inertia, $\tilde{r}$ is the average of the ellipsoid radii, $\mu$ the dynamic viscosity, $\vec{v}_{\|}$ the velocity parallel to the body, $V$ the body volume, $\mathbf{M}_A$ and $\mathbf{I}_A$ the diagonal virtual added-mass and virtual-added moment of inertia. Here $\left[ c_{blunt}, c_{slender}, c_{angular}, c_{kutta}, c_{magnus}\right]$ are the five fluid coefficients we identify. We ran the simulation on our computer with an Intel i9-11900K CPU, where the fish runs at 14.58x realtime, \textit{i.e.}, we can run \SI{14.58}{\second} of fish motions after \SI{1}{\second} of computation time.

As a baseline we compare with a classical, also stateless, formulation from Lighthill's Elongated Body Theory (EBT) model \cite{lighthill_note_1960}. This model for thrust and drag on a swimmer assumes that the lateral motion is small compared to the body length, which is valid for most undulating swimmers, including ours.

\begin{align}
\begin{split}
    T &= m \; \left[ \; \overline{\left( \frac{\partial h(t,x)}{\partial t} \right)^2} - U^2 \overline{\left( \frac{\partial h(t,x)}{\partial x} \right)^2} \; \right]_{x=L}
    \\
    D &= \frac{1}{2} \rho \; c_D \; S \; U^2
    \label{eq:EBT}
\end{split}
\end{align}
where $\overline{h(t,x)}$ implies the average lateral displacement over time, $U$ the cruising speed of the swimmer, $S$ is the total submerged surface area, $c_D$ is the drag coefficient, and the thrust expression is evaluated at the tip of the tail $x=L$. We consider the same local added mass per unit length from \cite{zhong_robot_2018} for $m = \frac{1}{4} \rho \pi \beta d^2$ evaluated at the tail tip with vertical submerged depth $d$ and a non-dimensional parameter $\beta$ that should be close to $1$. During constant cruising, which we consider for our data, the thrust and drag forces balance eachother, which results in a mapping from the tail oscillations $h(t,x)$ to the cruising speed $U$. We compute the spatial derivative of the lateral displacements $h(t, x)$, evaluated at the tip of the tail, using finite difference between the last two tail markers. The results can be found in \Cref{sec:comp-ebt-baseline}.



\section{System Identification}
\label{sec:sysid}

\subsection{Tail Stiffness}

\begin{figure}[!b]
    \centering
    \includegraphics[width=0.99\columnwidth]{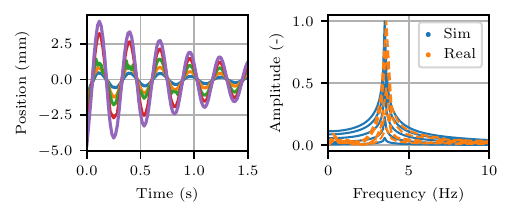}
    \caption{Natural frequency in both time and frequency domain of the fish tail measured on the real robot tail (left) and of the simulated system after matching the main amplitude frequency with reality (right). The different colors in the left plot indicate the 5 spine segments that are tracked during the oscillation.}
    \label{fig:fft}
\end{figure}

Since we have discretized the continuous tail as articulated rigid bodies, we identify the stiffness of the hinge joints based on the natural frequency of the system. We bend both the simulated and real system to a certain angle and release the tail to record a natural oscillation of the markers we track. We remove the tendons on both systems before conducting this experiment. Performing a Fast Fourier Transform (FFT) on both oscillations reveals a dominant frequency which we match by running a simple search over the scalar stiffness variable, which we assume to be the same value for all hinge joints in the discretized tail. The oscillations and frequencies can be seen in \Cref{fig:fft}. The dominant natural frequency we find is \SI{3.5}{\hertz}.

\begin{figure}[!t]
    \centering
    \includegraphics[width=0.99\columnwidth]{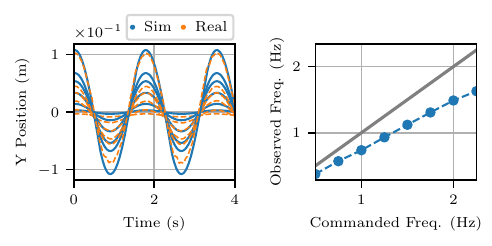}
    \caption{System identification of motor angular frequency and phase offset to minimize the distance between simulated and real markers, with a minimal distance of \SI{0.026}{\meter} for the \SI{0.60}{\hertz} swimming shown on the left.}
    \label{fig:OW}
\end{figure}

\subsection{Tail Actuation}

Since we have made several simplifications to the robot geometry, we perform another system identification to match the tail motion. We fix the head of the fish and track the markers of the spine segments outside of water while the motor is actuated at a constant angular velocity. By adjusting the motor arm length of the crank $c_m$ in simulation we try to match the actuated tail oscillations as best as possible. We observed that the real motor angular velocity is slightly slower than the commanded velocity, especially at the higher frequencies (rotations per second) likely due to reaching its torque limit, hence we additionally allow the simulated motor angular velocity $\omega$ and the phase offset $\phi$ to be optimized to match the real experiment. The phase offset is only needed because the sinusoidal movement of the tail, i.e. the motor position, is not guaranteed to start at the resting state in the experimental data. Since this is still a low dimensional search space we can sample the entire space to find an optimum that minimizes the mean distance of all markers between simulation and reality for a motion of a fixed \SI{4}{\second}. The minimization problem is defined as:
\begin{equation}
\min_{c_{m}, \omega, \phi} \; \frac{1}{T \cdot N}\sum_{t=1}^T\sum_{i=1}^N \left\| \mathbf{s}^i_{t}  - \overline{\mathbf{s}}^i_{t}\right \|_{2}
\label{eq:errOW}
\end{equation} 
where $\overline{\mathbf{s}}^i_{t}$ is the ground truth marker location $i$ at timestep $t$. We use a total of $N=11$ markers, of which $5$ belong to the rigid head and the remaining $6$ are the tail spine segments and the tail fin itself. The marker data, both simulated and real, is processed by rotating all markers into the local frame of the fish head, where the first marker on the head is defined as the origin and the vector from the first to the second marker is the positive x-axis. Each timestep we translate the local frame by the head displacement along the head vector of that timestep. The search space and the final result of the optimization is shown in \Cref{fig:OW}, where the markers for the head are not included in the visualization. 

We perform this out-of-water experiment on two angular velocities, one at a lower frequency of \SI{0.60}{\hertz} and another at a higher frequency of \SI{1.19}{\hertz}, of which the former is shown in \Cref{fig:OW}. The optimized distance errors for these two frequencies are \SI{26.0}{\milli\meter} and \SI{26.5}{\milli\meter} respectively. These optimized non-zero errors are mainly due to tracking inaccuracy, such as drifting markers, or even fabrication errors, where not all markers can be matched perfectly. This experiment is used to inform us on the motor arm length, which we choose to be \SI{0.0395}{\meter} as an average of the two optimization results, close to the measured real motor arm at \SI{0.04}{\meter}. For any following experiment, we always optimize the angular velocity of the motor in the real data to find the ``real'' frequency to instruct our simulated motor.


\subsection{Fluid Coefficients}
\label{sec:fluid}

Once we have found the motor arm length and angular velocity of swimming outside of water, we conduct sim-to-real for swimming experiments inside of water, including the fluid model in our simulated environment. Following the fluid model from \cite{vaxenburg2025whole}, there are five fluid coefficients to optimize: blunt drag, slender drag, angular drag, Kutta lift, and Magnus lift. Since this search space is nonconvex and higher dimensional, we combine a Bayesian Optimization method \cite{garrido2020dealing} for global search with a Nelder-Mead algorithm \cite{gao_implementing_2012} for more local refinement. Our objective is the same as \Cref{eq:errOW}, but now we optimize the fluid coefficients as well as a phase offset of the actuation. We set a reasonable bound for the fluid coefficients of $\left[0, 10 \right]$.

\begin{figure}[!t]
    \centering
    \includegraphics[width=0.99\columnwidth]{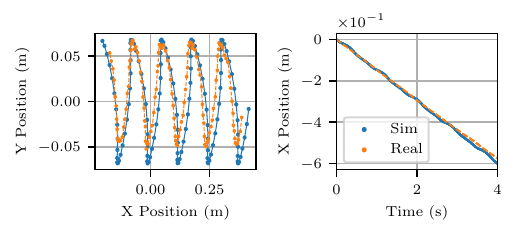}
    \caption{Optimized results of sim-to-real in-water experiment for \SI{1.19}{\hertz} data. We show the tail fin marker position (left) and forward swimming position (right) to validate the thrust generated by the hydrodynamics model. The average distance error on all markers is computed to be \SI{0.016}{\meter}.}
    \label{fig:IW}
\end{figure}

\begin{figure*}[!t]
    \centering
    \includegraphics[width=\textwidth]{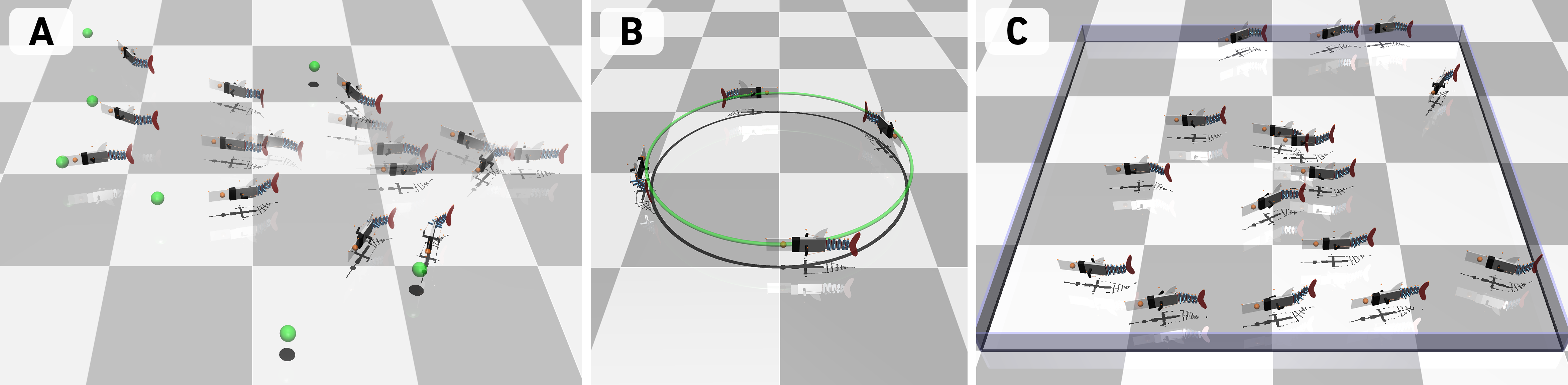}
    \caption{\textit{A)} SAC agent trained to track random target locations (green) with a $93\%$ success rate. Robot states are overlaid \SI{1}{\second} apart. \textit{B)} SAC agent tracking waypoints along the green circle trajectory with an average target error of \SI{0.009}{\meter}. \textit{C)} We demonstrate the ability to simulate a large number of swimmers in real-time.}
    \label{fig:examples}
\end{figure*}

We collected two sets of swimming trajectories on the hardware with constant tail oscillation frequency, the same \SI{0.60}{\hertz} and \SI{1.19}{\hertz} as before outside of water. The fluid coefficients were optimized using these two trajectories. For both the real and simulated data, we let the swimmer warmup until they reach their cruising state before we try to match simulation and reality, we choose this warmup time to be \SI{1}{\second}. The optimization result for the higher frequency can be seen in \Cref{fig:IW} where we observe an accurate forward swimming velocity throughout the entire trajectory with a final average distance error of \SI{28.8}{\milli\meter} for \SI{0.60}{\hertz} and \SI{16.1}{\milli\meter} for \SI{1.19}{\hertz}. We find the fluid coefficients $\left[ 0.40, 7.79, 2.81, 3.84, 0.27 \right]$. The full simulation pipeline with a single swimmer runs at $15\times$ real-time, and could be sped up more through parallelization. We showcase how a large number of robots can be simulated together in real-time in \Cref{fig:examples}.

\subsection{Comparison to EBT Baseline}
\label{sec:comp-ebt-baseline}
We can similarly optimize the drag coefficient in the EBT model to match our 2 experimental trajectories at \SI{0.60}{\hertz} and \SI{1.19}{\hertz}. We compute $\frac{\partial h}{\partial t}$ and $\frac{\partial h}{\partial x}$ through finite difference from the experimental marker data, and tune the $\beta$ in added mass per unit length and $c_D$ from \Cref{eq:EBT} such that the cruising speed $U$ matches our experimental observation. We use the same Bayesian Optimization as for the fluid coefficients in \Cref{sec:fluid}, with bounds of $\beta \in \left[0.9, 1.1 \right]$ and $c_D \in \left[0, 10 \right]$, and found optimal values $\beta=0.91$ and $c_D=0.31$. 

Note that the inherent nature of the optimization task for the fluid coefficients and EBT are different, since EBT optimizes directly to match cruising speed velocity, which is the target metric we care about, whereas our digital twin simulates the whole motion to match the real swimmer movement and extracts the forward velocity from this trajectory. Hence EBT can be seen more as a regression model on holistic descriptions of the swimmer, while the fluid coefficients describe the physics of the problem on a more detailed level of motion. This makes EBT less suitable for complex motions with controllable motor inputs, such as those encountered in reinforcement learning.

\section{Results}

\subsection{Sim-to-Real for Forward Swimming}

\begin{figure}[!b]
    \centering
    \includegraphics[width=0.99\columnwidth]{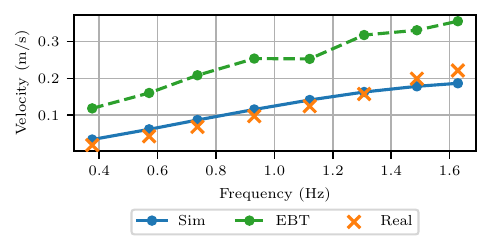}
    \caption{Sim-to-real generalization across frequencies of our optimized fluid coefficient model, \SI{0.019}{\meter\per\second} error, and the EBT model, \SI{0.134}{\meter\per\second}. Velocity direction is defined in the local fish frame.}
    \label{fig:freqs}
\end{figure}

With the previously identified robot and fluid parameters, we generalize the behavior to different swimming environments. We collected 8 more forward-swimming trajectories at different constant motor angular frequencies, and test the optimized model on its generalizability across swimming frequencies. As shown in \Cref{fig:freqs}, the simulated model can accurately capture the trend in forward swimming velocity of the robot, even though it was optimized on only two angular frequencies of swimming. The behavior at higher frequencies does show higher errors, with an average error in velocity of \SI{0.019}{\meter\per\second}. The physical robot could not reach higher actuation velocities due to the motor used. Since we optimize the observed motor frequency in the video and do not use the commanded actuation frequency, we show in \Cref{fig:freqs} how this error grows at larger angular velocities.

Comparing with the EBT baseline on the experimental frequency sweep data, we observe a significantly larger error for EBT in \Cref{fig:freqs} with an average error of \SI{0.134}{\meter\per\second}. We attribute this larger error to the marker data being potentially different enough that the computation of $\frac{\partial h}{\partial t}$ and $\frac{\partial h}{\partial x}$ for EBT is significantly affected between the previous 2 trajectories and the new 8 trajectories in this task. However, our digital twin seems to be robust to these changes, and can generalize without much loss of accuracy. We verify this hypothesis by optimizing the EBT parameters on our 8 new trajectories, and find a reduced error of \SI{0.027}{\meter\per\second}, which would not be a fair comparison, and is still significantly higher than the error using fluid coefficients.

\subsection{Reinforcement Learning for Target Tracking}

\begin{figure}[!b]
  \begin{subfigure}[!t]{0.495\columnwidth}
  \centering
    \includegraphics[width=\columnwidth]{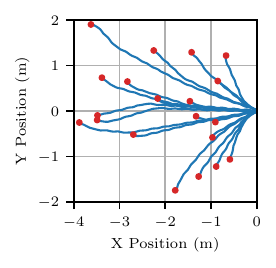}
  \end{subfigure}
  \hfill
  \begin{subfigure}[!t]{0.495\columnwidth}
  \centering
    \includegraphics[width=\columnwidth]{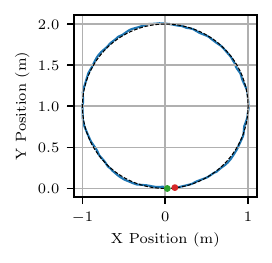}
  \end{subfigure}
  \caption{(Left) Tracking random waypoints in $20$ evaluation environments. The red points are the randomized target locations, while the blue curves are the trajectories of the swimmer. (Right) Tracking waypoints along circle with an average distance error of \SI{0.009}{\meter}. The green and red points indicate start and end positions respectively.}
  \label{fig:traj}
\end{figure}

Since our matched simulation environment can serve as a digital twin for learning strategies, we showcase this using a Soft-Actor-Critic (SAC) \cite{haarnoja_soft_2018} reinforcement learning algorithm trained with the objective to reach target positions. We set up the training environment in Stable Baselines 3 \cite{stable-baselines3}, and randomize the target location $\vec{x}_{target}$ in a $\SI{4}{\meter} \times \SI{4}{\meter}$ square in front of the fish head, an example for a distribution of target points can be seen in \Cref{fig:traj}.

\begin{table}[h]
\vspace{-1em}
    \centering
    \setlength{\tabcolsep}{7.5 pt}
    \renewcommand{\arraystretch}{1.4}
    \begin{tabular}[b]{l c}
        & \textbf{Observations} \\ 
        \toprule
        Motor Joint & $\cos(\alpha)$, $\sin(\alpha)$, $\dot\alpha$ \\
        Head Frame & $\dot\theta_x, \dot\theta_y, \dot\theta_z$ \\
        Tail Joint $i$ & $\phi_i, \dot\phi_i$ \\    
        Target Vector & $\| \vec{d}_{t} \|$, $\vec{d}_{t}$, $\dot{\vec{d}}_{t}$ \\
        Prev Action & $u_{old}$ \\
        \bottomrule
    \end{tabular}
    \caption{The observations used for training the SAC agent for target tracking. All observations are defined relative to the robot frame.}
    \label{tab:obs}
\end{table}

Due to the tendons in the fish spine, velocity control in MuJoCo tended to be more numerically stable than torque control for the motor. Consequently, we chose our control inputs $u(t)$ to be scalar accelerations for the motor such that $v_{new} = v_{old} + c_{action} u(t) dt$, where we can tune the multiplication factor $c_{action}$ and the default action space is $\left[ -1, 1 \right]$. The reward at each simulation step is defined as 
\begin{align}
    r(t) = -\| \vec{x}(t) - \vec{x}_{target} \| - \lambda \; | u(t) | + r_{goal}
\end{align}
where we found weighing factor $\lambda$ worked best when set to 1, and $r_{goal}$ is defined as $+300$ when the center of mass $\vec{x}(t)$ is within $\SI{5}{\centi\meter}$ of the target location, otherwise $0$. The simulation environment is reset when the goal reward is given. We limited the maximal tail flapping frequency to \SI{5}{\hertz} to keep results physically relevant. We defined all observations for the RL agent relative to the current position and orientation of the fish, where our observation space is $22$-dimensional, also seen in \Cref{tab:obs}, consisting of:
\begin{itemize}
    \item $\cos(\alpha)$, $\sin(\alpha)$, $\dot\alpha$: Motor joint position $\alpha$, choosing $\cos$ and $\sin$ to prevent discontinuities.
    \item $\dot\theta_x, \dot\theta_y, \dot\theta_z$: Angular velocities of head frame orientation.
    \item $\phi_i, \dot\phi_i$: Tail hinge joint $i$ position and velocity.
    \item $\| \vec{d}_{t} \|$, $\vec{d}_{t}$, $\dot{\vec{d}}_{t}$: Vector in local frame of the fish head pointing from the center of mass to the target.
    \item $u_{old}$: Previous action.
\end{itemize}

We ran a Bayesian hyperparameter optimization on the SAC training parameters, including learning rate, number of layers and layer dimension of the policy network, training frequency, when learning starts, soft update coefficient of SAC, and the action multiplier $c_{action}$. From the search we conclude that $c_{action}$ is ideal in the range of $\left[ 400, 500 \right]$, and a large number of transitions collected before start of training helps as well (around $500\;000$). 
We found the best performing control policy to have $93\%$ success rate on reaching the target, sampled in $100$ evaluation environments, some of which are visualized in \Cref{fig:traj}. The failing targets either tend to be at too high of a curvature to reach, or the target was reached a bit further than the tolerance, hence no environment reset occurs, and the robot keeps swimming and does not know how to turn back. The latter is likely solved with further training, while the former would require more complex controls and trajectory planning, beyond the scope of this paper.

We evaluate our target tracking policy on following waypoints on a \SI{1}{\meter} radius circle. The results in \Cref{fig:traj} show the tracking behavior, where the waypoint is moving over time along the trajectory. We compute the closest distance to the circular curve at each timestep, and find a mean distance of \SI{0.009}{\meter} with a maximum of \SI{0.023}{\meter}. Given that the RL agent is trained to reach within $\SI{5}{\centi\meter}$ of the target, this tracking performance of an average \SI{0.9}{\centi\meter} distance has already reached, if not exceeded, the best result it can produce. The results show that the simple, single degree-of-freedom velocity control of tail deflection can result in a wide variety of motions. The balance of simple hardware with complex control demonstrate that asymmetric actuation signals can be learned to cause turning, acceleration, and deceleration.

\section{Conclusion and Future Work}


In this paper we show the applicability of a simplified 5-parameter fluid model for bridging the sim-to-real gap on an underwater swimming fish robot. We show that with as little as 2 trajectories we can identify plausible fluid coefficients that can generalize to a range of angular motor frequencies for real-world swimming experiments, which simpler analytical models such as EBT struggled with. We demonstrate that this efficient digital twin can then be used for downstream tasks such as training RL agents to, for example, track target waypoints with a $93\%$ success rate. 

For further tasks such as shape optimization more expressive fluid models would be needed, since we believe stateless fluid descriptions cannot capture the intricacies of robot shape changes. One potential increase in complexity in the fluid model is tuning the fluid coefficients of each segment of the body separately, or use a more advanced extension to EBT such as \cite{boyer2010poincare}. Another limitation of the current work lies in the experimental approach of having a fully buoyant surface swimmer, instead of being able to explore a full underwater 3D space. This extension would require fully underwater state estimation and pitch or buoyancy control, which we will look into as a next step. We have considered constant cruising speed conditions for the sim-to-real in this paper, which opens up future challenges in turning and acceleration/deceleration behavior. 

Overall, this paper presents an initial step towards computationally efficient methods that can approximate soft robot swimmers in a realistic manner, better than previous analytical solutions. We show how these underwater swimmers can be easily included in standard pipelines for the robot learning community, and hope to reach a wider audience with these simple-to-build and simple-to-deploy biomimetic swimming robots.




\bibliographystyle{IEEEtran}
\bibliography{bibliography}

\end{document}